\documentclass[10pt,journal,compsoc]{IEEEtran}

\usepackage[cmex10]{amsmath}
\usepackage{acronym}
\usepackage{algorithmic}
\usepackage{array}
\usepackage{mdwmath}
\usepackage{mdwtab}
\usepackage{eqparbox}
\usepackage{fixltx2e}
\usepackage{stfloats}
\usepackage{url}
\usepackage{graphicx}
\usepackage[font=footnotesize,labelfont=bf]{subcaption}
\usepackage{hyperref}

\usepackage{multirow}

\usepackage[
backend=biber,
style=ieee,
]{biblatex}
\begin{document}
%
\title{Real Time Face Recognition System}
\vspace{-1.5cm}
\author{Adarsh Ghimire,
        Naoufel Werghi,
        Sajid Javed,
        Jorge Dias}

\markboth{Journal of \LaTeX\ Class Files,~Vol.~6, No.~1, January~2007}%
{Shell \MakeLowercase{\textit{et al.}}: Bare Advanced Demo of IEEEtran.cls for Journals}

\vspace{-1.5cm}
\IEEEcompsoctitleabstractindextext{%
\begin{abstract}
Over the past few decades, interest in algorithms for face recognition has been growing rapidly and has even surpassed human-level performance. Despite their accomplishments, their practical integration with a real-time performance-hungry system is not feasible due to high computational costs. So in this paper, we explore the recent, fast, and accurate face recognition system that can be easily integrated with real-time devices, and tested the algorithms on robot hardware platforms to confirm their robustness and speed.  
\end{abstract}

}
\vspace{-5cm}

\maketitle

\section{Introduction}
\vspace{-0.2cm}
\IEEEPARstart{F}{ace} is the most prominent biometric feature of the person and is widely used in finance, public security, and military etc, thus the researchers from computer vision domain, have heavily researched this topic for person identification. The state of the art deep face recognition system has already surpassed human level performance in most of the scenarios \cite{Deep_face_recognition_2021}. The employment of such face recognition systems with the robot system can provide robots with a capability to identify the person of interest in real-time. However, the key bottleneck in employing the latest face recognition system with robots lies in integrating the face detector with identification system. The face detection system which allows the system to identify where the faces are in the current frame, such that  identification module can confirm the identify of those particular faces, however face detection systems are quite heavy in themselves and thus requires more processing time. Even though there are lightweight traditional face detectors but they are not robust and do not perform well when the person’s faces in the frame are small or the frame environment is slightly dark \cite{A_survey_of_face_detection_2017}. However, there are deep learning based face detectors which when deployed along with GPU provide near real time and accurate results in complicated situations. It is necessary to clear the bottleneck lying ahead on the detection system, because the lag caused by the detector will affect the remaining part of the face recognition system. On the other hand, more accurate and robust face identification module are also heavy in themselves too, thus requires more computational power to perform accurate recognition. 

In this paper, we explored the recent, fast, and accurate face detection and recognition system to allow direct use of the face recognition system in robot platform. The system uses a yoloV5\cite{qi2021yolo5face} object detector trained on face dataset to detect face from the images and works at 152 FPS on Jetson TX2 board with some optimization, and mobile facenet \cite{chen2018mobilefacenets} based verification module that works at 75 FPS regardless of number of faces in the frames.

The main contributions of this paper are :\\ 
 1) \hspace{0.1cm} Fast and robust face detection and verification system\\
 2) \hspace{0.1cm} A real-time implementable face recognition system
\vspace{-0.3cm}

\section{Methods and Results}
\vspace{-0.1cm}
\subsection{Face Detection System}
\vspace{-0.1cm}
    Several popular face detection algorithms were explored initially Dlib, MTCNN, RetinaFace, yolov5 etc. The precision of the several algorithms evaluated on WiderFace dataset are reported in Table \ref{tab:face_detection_precision}. In addition to the performance metrics, the total number of parameter and total number of Floating point operations (FLOPS) required by different algorithms are also reported. For real-time operation in robot processor, one with the smaller FLOPS has be to selected. Thus, the models under the category of Large type were not considered for further evaluation since they were very heavy, required high computational power, and did not have significant performance difference. 
    
    Under the small category, the one with least FLOPS is FaceBoxes, however the precision of it in easy set is less so it is not suitable because robustness cannot not be achieved. The selection between SCRFD-0.5GF and Yolo-V5n0.5\cite{qi2021yolo5face} is driven by the latest algorithm and easy compatibility with the system. Even though, SCRFD-0.5GF has slightly less FLOPS then Yolo-V5n0.5, the later has better accuracy in easy and hard set. In addition to that, the Yolo-V5 was written in pytorch thus it is easily integrable in the robot system. Selection between Yolo-V5n and Yolo-V5n0.5 is motivated by the number of FLOPS required and the accuracy. In the expense of 4 times more FLOPS, the Yolo-V5n can give only slightly higher precision compared to other. Since, the accuracy is already high so the FLOPS requirement was a significant factor to decide the choice of algorithm. Thus, Yolo-V5n0.5 was chosen for further evaluation and testing. 
    
\begin{table}[]
\caption{Face detection algorithms evaluation on WiderFace Dataset}
\label{tab:face_detection_precision}
\begin{tabular}{cccccc}
\hline
\multirow{2}{*}{\textbf{Model Type}} & \multirow{2}{*}{\textbf{Detector}} & \multicolumn{2}{c}{\textbf{Performance}} & \multirow{2}{*}{\textbf{Parameters(M)}} & \multirow{2}{*}{\textbf{Flops(G)}} \\
 &  & Easy & Hard &  &  \\ \hline
\multirow{9}{*}{Large} & MTCNN & 85.1 & 60.7 & - & - \\
 & DSFD & 94.3 & 71.4 & 120.06 & 259.55 \\
 & TinaFace & 95.6 & 81.4 & 37.98 & 172.95 \\
 & SCRFD-34GF & 96.1 & 85.3 & 9.8 & 34.13 \\
 & Yolo-V5l & 95.9 & 84.5 & 46.63 & 41.61 \\ \hline
\multirow{6}{*}{Small} & SCRFD-2.5GF & 93.8 & 77.9 & 0.67 & 2.53 \\
 & SCRFD-0.5GF & 90.6 & 68.5 & 0.57 & 0.51 \\
 & RetinaFace & 87.8 & 47.3 & 0.44 & 0.80 \\
 & FaceBoxes & 76.2 & 24.2 & 1.01 & 0.28 \\
 & Yolo-V5n & 93.6 & 80.5 & 1.73 & 2.11 \\
 & Yolo-V5n0.5 & 90.8 & 73.8 & 0.44 & 0.57 \\ \hline
\end{tabular}
\vspace{-0.3cm}
\end{table}
    
    From above exploration, the yolo-V5n0.5 was the chosen algorithm for further evaluation and will be referred as Yolo-V5. For the completeness, several popularly used fast face detection algorithms were tested on different platforms. Table \ref{tab:face_detection_speed} reports the speed of several face detection algorithms tested on CPU(8 Core), Colab, and Jetson TX2 Board. Started with testing of the detection algorithms on the local computer in order to solve the code issues. Then the algorithms were tested on Colab GPU to check the compatibility with Nvidia GPUs. Finally, once the algorithms successfully shows the enhanced performance on GPU then the Jetson board was set up to work on those algorithms. From Table \ref{tab:face_detection_speed} we can see that Yolo-V5 has 48 FPS on octa-core CPU, 175 FPS on Colab GPU, and only 17 FPS on Jetson Board. Eventhough Jetson has 256 core the performance is not enhanced, this is because the pytorch code are not optimized for Jetson. Thus, in order to solve this problem of low speed, the pytorch model was converted to TensorRT Engine file. Table \ref{tab:face_detection_speed} reports the Yolo-V5(TensorRT) detection speed of 152 FPS, which is around 10 times more than the original pytorch one.
    
    \begin{table}
        \caption{Speed performance comparision of face detection algorithms}
        \label{tab:face_detection_speed}
        \begin{tabular}{clc}
            \hline
            Platform & Algorithms & Speed(fps) \\ \hline
            \multirow{5}{*}{CPU} & DLib(CNN) & 2 \\
             & DLib(HOG) & 8 \\
             & MTCNN & 12 \\
             & Haar Cascade & 18.5 \\
             & Yolo-V5 (Pytorch) & 48 \\ \hline
            \multirow{3}{*}{Colab} & MTCNN & 30 \\
             & DLib(CNN) & 36 \\
             & Yolo-V5 (Pytorch) & 175 \\ \hline
            \multirow{3}{*}{Jetson} & MTCNN & 9 \\
             & Yolo-V5 (Pytorch) & 17 \\
             & \multicolumn{1}{c}{Yolo-V5(TensorRT)} & 152 \\ \hline
        \end{tabular}
        \vspace{-0.3cm}
    \end{table}
\vspace{-0.5cm}

\subsection{Face Verification System}
    For verification of the detected face in the scene, several face verification algorithms like facenet, mobile facenet, deep face etc were explored. The performance of different methods and their corresponding Large Faces in Wild (LFW) dataset accuracy are reported in Table \ref{tab:face_verification}. The table also reports about the size of the model, FLOPS, and total parameters in the model. Thus, after comparing several models and by taking under consideration the speed requirement and accuracy, the MobileFaceNet is chosen for further evaluation.
    
\begin{figure}[!t]
        \centering
        \includegraphics[scale =0.35]{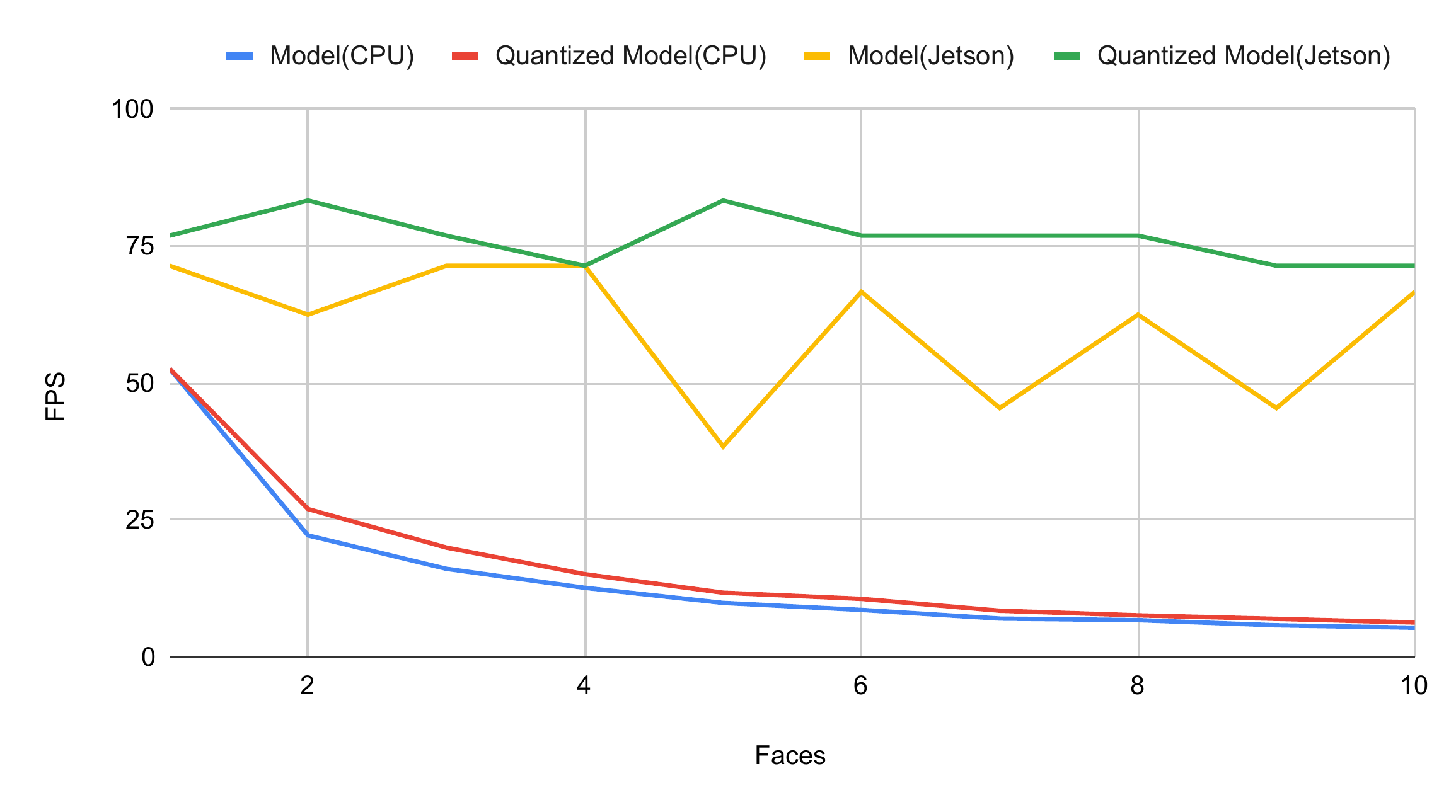}
        \vspace{-0.5cm}
        \caption{Mobile Facenet System Performance Chart}
        \label{fig:mobile_facenet_performance}
        \vspace{-0.7cm}
    \end{figure}

    The MobileFaceNet was further tested for real-time performance on CPU as well as on Jetson board, the result is shown in Figure \ref{fig:mobile_facenet_performance}. First, the model was tested on CPU then directly tested on Jetson. And, found that as the number of faces increased in the scene the performance on the CPU decreased (as shown by blue curve) however the performance on Jetson board fluctuated between 70 to 40 FPS (as shown by yellow curve). For further performance enhancement, the model weights were quantized to lower precision and further tested on CPU and Jetson board. As a result of quantization the performance on CPU remained similar to unquantized version. However, the performance on Jetson board for increasing number of faces resulted in stable performance. Meaning, as the number of faces increased the performance of the model remained at 75 FPS. This clears, that the model is ready for real-time application.
    \begin{table}[!t]
        \caption{Performance Comparision of Face verification methods on LFW dataset}
        \label{tab:face_verification}
        \begin{tabular}{ccccc}
        \hline
        Method & Accuracy & Parameters(M) & FLOPS & Model Size \\ \hline
        Facenet & 99.63 & 7.5 & 1.6B & 30 MB \\
        VGG-Face & 98.78 & 138 & 15G & - \\
        Light CNN-29 & 99.33 & 12.6 & 3.9G & 125 MB \\
        ShuffleFaceNet & 99.67 & 2.6 & 577.5M & 10.5 MB \\
        MobileFaceNet & 99.55 & 1 & 439.8M & 4 MB \\ \hline
        \end{tabular}
    \end{table}
    
\section{Conclusion}
In this work, we found that YoloV5 with optimization using TensorRT can achieve ten times faster speed in Jetson TX2 than the normal PyTorch model. Using a mobile-facenet for face verification can provide up to 75 FPS regardless of the number of faces in the given frame. This concludes that by combining yoloV5 with mobile-facenet we can achieve a highly robust and fast face recognition system.
\printbibliography
\end{document}